\documentclass{article}
\usepackage{iclr2022_workshop,times}

\usepackage{amsmath,amsfonts,bm}

\def\eqref#1{equation~\ref{#1}}

\def\1{\bm{1}}

\DeclareMathAlphabet{\mathsfit}{\encodingdefault}{\sfdefault}{m}{sl}
\SetMathAlphabet{\mathsfit}{bold}{\encodingdefault}{\sfdefault}{bx}{n}

\DeclareMathOperator*{\argmax}{arg\,max}

\usepackage[utf8]{inputenc} 
\usepackage[T1]{fontenc}    
\usepackage{hyperref}       
\usepackage{url}            
\usepackage{booktabs}       
\usepackage{amsfonts}       
\usepackage{nicefrac}       
\usepackage{microtype}      
\usepackage{xcolor}         
\usepackage{amsmath}
\usepackage{graphicx}
\usepackage{comment}
\usepackage[linesnumbered, ruled]{algorithm2e}

\SetCommentSty{commentfont}

\graphicspath{{figures/}}

\title{AdaGrid: Adaptive Grid Search\\for Link Prediction Training Objective}

\author{%
  Tim Poštuvan\\
  EPFL\\
  \texttt{tim.postuvan@epfl.ch} \\
  \And
  Jiaxuan You \\
  Stanford University \\
  \texttt{jiaxuan@cs.stanford.edu} \\
  \And
  Mohammadreza Banaei \\
  EPFL \\
  \texttt{mohammadreza.banaei@epfl.ch} \\
  \And
  Rémi Lebret \\
  EPFL \\
  \texttt{remi.lebret@epfl.ch \hspace*{0.62cm}} \\
  \And
  Jure Leskovec \\
  Stanford University \\
  \texttt{jure@cs.stanford.edu}
}

\iclrfinalcopy 

\begin{document}

\maketitle

\begin{abstract}

One of the most important factors that contribute to the success of a machine learning model is a good training objective. Training objective crucially influences the model’s performance and generalization capabilities. This paper specifically focuses on graph neural network training objective for link prediction, which has not been explored in the existing literature. Here, the training objective includes, among others, a negative sampling strategy, and various hyperparameters, such as edge message ratio which controls how training edges are used. Commonly, these hyperparameters are fine-tuned by complete grid search, which is very time-consuming and model-dependent. To mitigate these limitations, we propose Adaptive Grid Search (AdaGrid), which dynamically adjusts the edge message ratio during training. It is model agnostic and highly scalable with a fully customizable computational budget. Through extensive experiments, we show that AdaGrid can boost the performance of the models up to $1.9\%$ while being nine times more time-efficient than a complete search. Overall, AdaGrid represents an effective automated algorithm for designing machine learning training objectives. Code is available at \url{https://github.com/timpostuvan/adagrid}.
\end{abstract}

\section{Introduction}

Link prediction is one of the most important tasks on graph-structured data. For a given pair of entities, the goal of link prediction is predicting whether they are going to interact. Applications of link prediction are found in various fields, such as social networks, recommender systems, and biology. There have been many strategies to cope with the link prediction task \citep{Lu2011}, where the state-of-the-art approaches use Graph Neural Networks (GNN) \citep{Kipf2016a, Hamilton2017, Zhou2020}. \citet{Kipf2016b} introduced a variational graph autoencoder (VGAE), which embeds nodes in latent space so that the graph's adjacency matrix can be reconstructed from their representations. \citet{Zhang2017} proposed a novel framework called Weisfeiler-Lehman Neural Machine (WLNM), which is based on the Weisfeiler-Lehman algorithm \citep{shervashidze2011}. Furthermore, \citet{Zhang2018b} introduced SEAL, a new heuristic learning paradigm, which captures first, second, and higher-order structural information in the form of local subgraphs.

To train GNNs, an appropriate training objective has to be selected, which includes choice of the objective function, evaluation metric, and training strategy. In this paper, we specifically focus on how to best exploit training data by altering the edge message ratio parameter (i.e., the proportion of edges used for message passing and loss calculation). This problem has not been explored yet in the existing literature despite it crucially influencing the performance of a model and its generalization capabilities. The standard approach to fine-tune the link prediction objective is a complete search over a certain hyperparameter space \citep{zhou2019auto, you2020design}. Complete search trains the model multiple times for the full number of epochs, each time with a different edge message ratio from a set of predefined values, and selects the one with the maximal final validation AUC. Even though it exhaustively searches the hyperparameter space, which is very time-consuming, it still obtains non-optimal performance. Moreover, even when the configuration of the model slightly changes, a complete search has to be repeated.

In this paper, we propose Adaptive Grid Search (AdaGrid) that adjusts edge message ratio during training to each specific model and dataset, therefore alleviating time complexity and suboptimal performance of the complete search. AdaGrid is model agnostic and highly scalable with a fully customizable computational budget. Through extensive experiments, we show that AdaGrid can boost the performance of the models up to $1.9\%$ while being nine times more efficient than a complete search. We also propose a new negative sampling strategy, which samples harder negative instances than the standard uniform negative sampling by considering the community structure of networks.

\section{Proposed Method: Adaptive Grid Search (AdaGrid)}
\label{adaptive-grid-search-adagrid}

\subsection{Standard Learning-based Link Prediction Experimental Setting}

Learning-based link prediction task is often formulated as binary classification, where potential edges are classified as true or false. The standard learning-based link prediction setting \citep{Zhang2018b} first splits the graph's edge set into a training set, validation set, and test set according to the split ratio (i.e., training/validation/test split ratio). For link prediction using GNN models, each set of edges has to be further divided into message-passing edges and objective edges. The message-passing edges are used for propagating information between nodes, while the loss is calculated based on the objective edges. So far, objective edges comprise only positive instances; therefore, negative instances (i.e., nonexistent edges) are also added to the objective edge sets by uniformly random sampling node pairs for a balanced classification task. In this paper, we focus on the setting where message-passing and objective edge sets are disjoint, and \textit{edge message ratio} hyperparameter controls the proportion of message-passing edges. Furthermore, every fixed number of epochs both sets are resampled so training can take full advantage of all edges for both purposes.

\subsection{Complete Search}
\label{complete-search}

To find a good training objective, the standard approach to setting a good edge message ratio is a complete search over some predefined set of values $\mathcal{Q}$ with cardinality $L = |\mathcal{Q}|$ \citep{you2020design}. Complete search is very time-consuming and computationally demanding because the model has to be trained multiple times with different edge message ratios. A possible speedup is to train each version of the model only for part of all epochs and then take the one which performs best at that time. While this seems like an adequate solution, there are still certain limitations. Even by slightly changing the configuration of the model, the optimal edge message ratio changes, and the complete search has to be repeated, which is extremely inconvenient. Another, possibly false, assumption of this approach is that the optimal edge message ratio does not change during training. To resolve these drawbacks of the standard approach, we propose Adaptive Grid Search (AdaGrid).

\subsection{Adaptive Grid Search (AdaGrid)}

The key feature of AdaGrid is its ability to adapt edge message ratio during training to each configuration of the model and each dataset. AdaGrid is described in Algorithm \ref{algorithm-AdaGride}. It changes the edge message ratio every \textit{adapt epochs} $\alpha$. Then, it trains $L$ copies of the model with different edge message ratios from a set of predefined edge message ratios $\mathcal{Q}$ in parallel, where each copy is trained for \textit{try epochs} $\beta \leq \alpha$. After \textit{try epochs} of training, the new edge message ratio is selected based on one of two possible criteria: \textit{validation criterion} and \textit{gap criterion}. The validation criterion selects the edge message ratio corresponding to the model with the highest final validation AUC. On the other hand, the gap criterion chooses the edge message ratio of the model with minimal absolute difference between the final training and validation AUC. Since sometimes training and validation AUCs are a bit unstable during training, additionally smoothing can be performed (i.e., instead of the final training/validation AUC, rather average of a few last training/validation AUCs is taken). One of the main upsides of AdaGrid is its flexibility since it can be adjusted to each application separately. The adapt epochs $\alpha$ and the try epochs $\beta$ regulate training time, while the selection criterion and set of considered edge message ratios $\mathcal{Q}$ can be tailored to each specific task.

\paragraph{Computational budget of AdaGrid} 
By setting the adapt epochs $\alpha$ and the try epochs $\beta$ appropriately, AdaGrid's computational budget can be fully customizable. If the model is trained for number of epochs $N$ and $L$ edge message ratios are considered, the overall number of training epochs of AdaGrid is:
\begin{equation}
	epochs_{AdaGrid} = N \cdot \left( 1 + \frac{(L - 1) \cdot \beta}{\alpha} \right),
\end{equation}
\noindent
while complete search requires:
\begin{equation}
	epochs_{complete\; search} = N \cdot L.
\end{equation}
\noindent
For instance, if try epochs and adapt epochs are equal: $\beta = \alpha$, both required the same number of epochs, while if $\beta = 5$, $\alpha = 50$, and $L = 9$, AdaGrid needs five times fewer training epochs.

\begin{algorithm}[h]
	\caption{AdaGrid}
	\label{algorithm-AdaGride}
	\DontPrintSemicolon
	
	\SetKwInOut{Input}{Input}
	\SetKwInOut{Output}{Output}

    \Input{Dataset $\mathcal{D}$, model weights $\theta_{0}$, \\
           number of epochs $N$, adapt epochs $\alpha$, try epochs $\beta \leq \alpha$,  \\
           set of considered edge message ratios $\mathcal{Q}$, \\
           training function \textsc{train}, selection criterion $f$ \\ 
	}
	\Output{Trained model weights $\theta_{n}$
	}
	
	\BlankLine
	$n \gets \frac{N}{\alpha}$ \;
	\For{$i \gets 1$ \KwTo $n$}{
	    \For{$q \in \mathcal{Q}$}{
	        $\theta_{i, q}, AUC^{train}_{i, q}$, $AUC^{val}_{i, q} \gets$ \textsc{train}($\mathcal{D}$, $\theta_{i-1}$, $\beta$, $q$) \;
	    }
	    
	    $q_{opt}\gets \argmax_{q} f(AUC^{train}_{i, q}$, $AUC^{val}_{i, q})$ \;
	    $\theta_{i},$ $AUC^{train}_{i},$ $AUC^{val}_{i} \gets $ \textsc{train}($\mathcal{D}$, $\theta_{i, q_{opt}}$, $\alpha - \beta$, $q_{opt}$) \;
	}
	\Return $\theta_{n}$ \;
\end{algorithm}

\section{Experimental Results}
\label{experiments}

We test the performance of AdaGrid on various model configurations, datasets, data split ratios, negative sampling strategies, and hyperparameter settings of AdaGrid. All experiments are conducted on Cora and CiteSeer datasets \citep{Sen2008}. Both datasets are well-known citation networks (more details in Appendix \ref{dataset-details}). Model configuration and experimental set-up are described in Appendix \ref{model-configuration-and-experimental-set-up}.

\paragraph{Baselines}
To contextualize the empirical results of AdaGrid, we compare it against two baselines: complete search and random search. Complete search (see Section \ref{complete-search}) exhaustively searches hyperparameter space, which makes it very time-consuming. It also does not change the edge message ratio during training. On the other hand, random search modifies the edge message ratio to a random value from $[0.1, 0.9]$ interval after every training epoch. This makes it very fast, however, different edge message ratios are not inspected. We show that both baselines perform inferior to AdaGrid: complete search has static edge message ratio and is slow, while random search does not explore edge message ratio space.

\paragraph{Community ratio-based negative sampling}

Uniform negative sampling seems to be an appropriate approach for obtaining negative instances, however, it turns out that these instances are rather simple negative examples for link prediction. A lot of networks inherently display some kind of community structure (i.e., the network can be partitioned into disjoint communities so that connections within communities are denser than the connections with the rest of the network) \citep{fang2020survey}. Let us define edges that have both endpoints in the same community as the \emph{within community edges}, and edges that have endpoints in the different communities as the \emph{between communities edges}. Furthermore, let \textit{community ratio} be the proportion of node pairs with nodes from the same community. Uniform negative sampling yields mainly between community edges, which creates an easy evaluation setting. To make a harder evaluation setting, we propose community ratio-based negative sampling that obtains negative edges in such a way that sets of negative and positive instances can not be distinguished based on community ratio. It first performs community detection on the graph with all training edges, then it measures the community ratio on validation edges. Afterward, negative instances for all three sets are sampled in compliance with the gauged community ratio. Our approach is beneficial because it generates more challenging negative instances, which can better differentiate the performance of models even on easier datasets.

\subsection{AdaGrid and Uniform Negative Sampling}
\label{adagrid-and-uniform-negative-sampling}

We first evaluate AdaGrid on a uniform negative sampling setting because this is the standard approach for link prediction. Table \ref{table-adagrid-uniform-negative-sampling} shows that AdaGrid consistently performs better than the best baseline approach, no matter its configuration. Improvement of more than $0.7\%$ is especially evident for the $80/10/10$ split ratio, which is the most similar to the usual experimental data splits. Another crucial aspect of AdaGrid is its adjustability in terms of computational budget. Even when AdaGrid is trained for considerably fewer training epochs than the complete search, it constantly performs better than the best baseline approach. Table \ref{table-adagrid-uniform-negative-sampling-time} exhibits that AdaGrid can surpass complete search and random search for more than $0.7\%$ even when it requires five times fewer epochs than a complete search and fewer than twice as many as a random search ($\alpha=10$ and $\beta=1$). AdaGrid even outperforms both of them while being nine times faster than a complete search and is computationally almost equivalent to the training of a single model ($\alpha=100$ and $\beta=1$).


\begin{table}[h]
    \caption{AUC in percent for AdaGrid with adapt epochs $\alpha = 10$ and try epochs $\beta = 1$, complete search, and random search with uniform negative sampling evaluation.}
    \label{table-adagrid-uniform-negative-sampling}
    \centering
    \resizebox{\columnwidth}{!}{
    \begin{tabular}{ccccccc}
        \toprule
        & \multicolumn{6}{c}{\textbf{Datasets}}           \\
        & \multicolumn{3}{c}{Cora} & \multicolumn{3}{c}{CiteSeer} \\
        \cmidrule(lr){2-4} \cmidrule(lr){5-7}
        \textbf{Methods} & $20/40/40$ & $50/25/25$ & $80/10/10$ & $20/40/40$ & $50/25/25$ & $80/10/10$ \\
        \midrule
        Complete search & 94.85 & 96.65 & 97.17 & 95.84 & 97.78 & 98.46 \\
        Random search & 94.71 & 96.40 & 97.02 & \textbf{95.88} & 97.67 & 98.45 \\
        \textbf{AdaGrid} & \textbf{95.01} & \textbf{97.07} & \textbf{97.90} & 95.79 & \textbf{97.89} & \textbf{98.70} \\
        \midrule
        Gain & 0.16 & 0.42 & 0.73 & -0.09 & 0.11 & 0.24 \\
        \bottomrule
    \end{tabular}
    }
\end{table}

\begin{table}[h]
  \caption{AUC in percent and relative number of epochs for different configurations of AdaGrid, complete search, and random search with with uniform negative sampling evaluation.}
  \label{table-adagrid-uniform-negative-sampling-time}
  \centering
  \resizebox{\columnwidth}{!}{
  \begin{tabular}{cccccccccccc}
    \toprule
    & \multicolumn{2}{c}{\textbf{Baselines}} & \multicolumn{9}{c}{\textbf{AdaGrid (adapt epochs $\alpha$ /try epochs $\beta$)}} \\
    & Random & Complete & 100/1 & 100/5 & 100/100 & 50/1 & 50/5 & 50/50 & 10/1 & 10/5 & 10/10 \\ 
    \midrule
    AUC & 97.02 & 97.17 & 97.54 & 97.63 & 97.59 & 97.68 & 97.75 & 97.76 & \textbf{97.90} & 97.83 & 97.86 \\
    Epochs & 1.00 & 9.00 & 1.08 & 1.40 & 9.00 & 1.16 & 1.80 & 9.00 & 1.80 & 5.00 & 9.00 \\
    \bottomrule
  \end{tabular}
  }
\end{table}

\subsection{AdaGrid and Community Ratio-based Negative Sampling}
\label{adagrid-and-community-ratio-based-negative-sampling}

We also evaluate AdaGrid on the proposed community ratio-based negative sampling setting to display its advantages, as well as advantages of AdaGrid. Table \ref{table-adagrid-community-ratio-based-negative-sampling} shows that AdaGrid nearly always performs better than the best baseline approach, regardless of its configuration. Comparing results of negative sampling strategies in Tables \ref{table-adagrid-uniform-negative-sampling} and \ref{table-adagrid-community-ratio-based-negative-sampling} shows that community ratio-based negative sampling creates a more challenging evaluation setting since AUC scores are considerably lower. It also better differentiates the performance of models since gains of AdaGrid are always bigger than the ones of uniform negative sampling. Especially noteworthy is the gain of more than $1.9\%$, which again confirms the benefits of AdaGrid.

\begin{table}[h]
    \caption{AUC in percent for AdaGrid with adapt epochs $\alpha = 10$ and try epochs $\beta = 5$, complete search, and random search with community ratio-based negative sampling evaluation.}
    \label{table-adagrid-community-ratio-based-negative-sampling}
    \centering
    \resizebox{\columnwidth}{!}{
    \begin{tabular}{ccccccc}
        \toprule
        & \multicolumn{6}{c}{\textbf{Datasets}}           \\
        & \multicolumn{3}{c}{Cora} & \multicolumn{3}{c}{CiteSeer} \\
        \cmidrule(lr){2-4} \cmidrule(lr){5-7}
        \textbf{Methods} & $20/40/40$ & $50/25/25$ & $80/10/10$ & $20/40/40$ & $50/25/25$ & $80/10/10$ \\
        \midrule
        Complete search & 84.01 & 82.98 & 84.62 & \textbf{83.87} & 82.22 & 83.96  \\
        Random search & 84.00 & 82.05 & 83.82 & \textbf{83.87} & 82.61 & 83.65 \\
        \textbf{AdaGrid} & \textbf{84.65} & \textbf{83.82} & \textbf{86.54} & 83.75 & \textbf{83.37} & \textbf{84.89} \\
        \midrule
        Gain & 0.64 & 0.84 & 1.92 & -0.12 & 0.76 & 0.93 \\
        \bottomrule
    \end{tabular}
    }
\end{table}

\subsection{AdaGrid and Edge Message Ratio}
\label{adagrid-and-edge-message-ratio}

\begin{figure}[h]
	\centering
	\includegraphics[width=\textwidth]{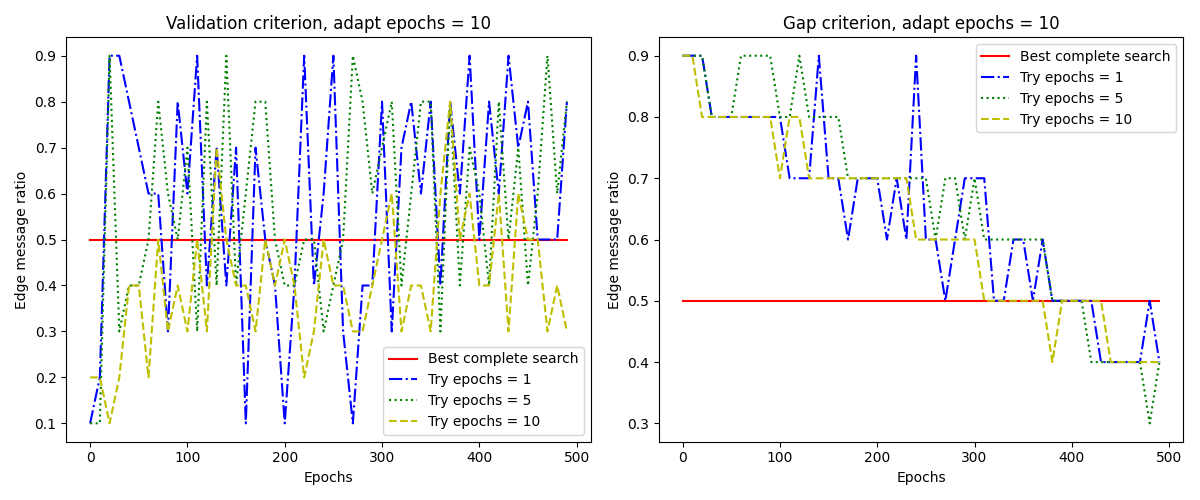}
	\caption{Edge message ratio during training, regulated by validation criterion and gap criterion.} 
	\label{fig-adagrid-edge-message-ratio}
\end{figure}

The success of AdaGrid probably stems from its ability to change edge message ratio during training in an informed way. According to Figure \ref{fig-adagrid-edge-message-ratio}, AdaGrid modifies edge message ratio almost every $\alpha$ adapt epochs in conjunction with both criteria. Additionally, Table \ref{table-adagrid-uniform-negative-sampling-time} shows that AdaGrid performs better with a lower number of adapt epochs $\alpha$ (i.e., it is beneficial to be capable of changing edge message ratio more frequently). Therefore, complete search results in non-optimal performance because it assumes the edge message ratio is static. On the other hand, random search does not explore edge message ratio space, so it incorrectly alters it.

\section{Conclusion}

This paper aims to explore graph neural network training objectives for link prediction. We expose problems of finding a good training objective and the current standard approach (i.e., complete search). To diminish the inconvenience of edge message ratio fine-tuning, we propose AdaGrid, which adapts edge message ratio ``on-the-fly'' during training and overcomes limitations of complete search. It is model agnostic and has a fully customizable computational budget. More importantly, AdaGrid can reduce training time and boost performance at once. It can improve the performance of the models up to $1.9\%$, while it can be nine times more efficient than a complete search. We also propose community-based negative sampling, which samples harder negative instances and creates a more challenging link prediction evaluation setting.

\paragraph{Future work}
Since AdaGrid performs well on link prediction, it would be interesting to apply AdaGrid to other graph learning tasks. Due to its generality, it would be suitable even for other deep learning tasks. Instead of edge message ratio, AdaGrid can optimize any parameter which can dynamically change during training. To further speed up AdaGrid, instead of considering all edge message ratios, only those which are adjacent to the current one can be considered during adaptation phases. Another intriguing idea is to eliminate the predefined set of values from which edge message ratios are selected. By locally interpolating validation AUCs near the current edge message ratio, a new edge message ratio can be chosen to maximize validation AUC.

\bibliography{references}

\begin{thebibliography}{14}
\providecommand{\natexlab}[1]{#1}
\providecommand{\url}[1]{\texttt{#1}}
\expandafter\ifx\csname urlstyle\endcsname\relax
  \providecommand{\doi}[1]{doi: #1}\else
  \providecommand{\doi}{doi: \begingroup \urlstyle{rm}\Url}\fi

\bibitem[Clauset et~al.(2004)Clauset, Newman, and Moore]{Clauset2004}
Aaron Clauset, M.~E.~J. Newman, and Cristopher Moore.
\newblock Finding community structure in very large networks.
\newblock \emph{Phys. Rev. E}, 70:\penalty0 066111, Dec 2004.
\newblock \doi{10.1103/PhysRevE.70.066111}.
\newblock URL \url{https://link.aps.org/doi/10.1103/PhysRevE.70.066111}.

\bibitem[Fang et~al.(2020)Fang, Huang, Qin, Zhang, Cheng, and
  Lin]{fang2020survey}
Yixiang Fang, Xin Huang, Lu~Qin, Wenjie Zhang, Reynold Cheng, and Xuemin Lin.
\newblock A survey of community search over big graphs.
\newblock \emph{The VLDB Journal}, 29, 01 2020.
\newblock \doi{10.1007/s00778-019-00556-x}.

\bibitem[Hamilton et~al.(2017)Hamilton, Ying, and Leskovec]{Hamilton2017}
William~L Hamilton, Rex Ying, and Jure Leskovec.
\newblock Inductive representation learning on large graphs.
\newblock In \emph{Proceedings of the 31st International Conference on Neural
  Information Processing Systems}, pp.\  1025--1035, 2017.

\bibitem[Kipf \& Welling(2016{\natexlab{a}})Kipf and Welling]{Kipf2016a}
Thomas~N Kipf and Max Welling.
\newblock Semi-supervised classification with graph convolutional networks.
\newblock \emph{arXiv preprint arXiv:1609.02907}, 2016{\natexlab{a}}.

\bibitem[Kipf \& Welling(2016{\natexlab{b}})Kipf and Welling]{Kipf2016b}
Thomas~N. Kipf and Max Welling.
\newblock Variational graph auto-encoders, 2016{\natexlab{b}}.

\bibitem[Loshchilov \& Hutter(2017)Loshchilov and Hutter]{Loshinov2017}
Ilya Loshchilov and Frank Hutter.
\newblock {SGDR:} stochastic gradient descent with warm restarts.
\newblock In \emph{5th International Conference on Learning Representations,
  {ICLR} 2017, Toulon, France, April 24-26, 2017, Conference Track
  Proceedings}. OpenReview.net, 2017.
\newblock URL \url{https://openreview.net/forum?id=Skq89Scxx}.

\bibitem[Lü \& Zhou(2011)Lü and Zhou]{Lu2011}
Linyuan Lü and Tao Zhou.
\newblock Link prediction in complex networks: A survey.
\newblock \emph{Physica A: Statistical Mechanics and its Applications},
  390\penalty0 (6):\penalty0 1150--1170, 2011.
\newblock ISSN 0378-4371.
\newblock \doi{https://doi.org/10.1016/j.physa.2010.11.027}.
\newblock URL
  \url{https://www.sciencedirect.com/science/article/pii/S037843711000991X}.

\bibitem[Sen et~al.(2008)Sen, Namata, Bilgic, Getoor, Gallagher, and
  Eliassi-Rad]{Sen2008}
P.~Sen, Galileo Namata, M.~Bilgic, L.~Getoor, B.~Gallagher, and Tina
  Eliassi-Rad.
\newblock Collective classification in network data.
\newblock \emph{AI Mag.}, 29:\penalty0 93--106, 2008.

\bibitem[Shervashidze et~al.(2011)Shervashidze, Schweitzer, van Leeuwen,
  Mehlhorn, and Borgwardt]{shervashidze2011}
Nino Shervashidze, Pascal Schweitzer, Erik~Jan van Leeuwen, Kurt Mehlhorn, and
  Karsten~M. Borgwardt.
\newblock Weisfeiler-lehman graph kernels.
\newblock \emph{J. Mach. Learn. Res.}, 12:\penalty0 2539--2561, 2011.

\bibitem[You et~al.(2020)You, Ying, and Leskovec]{you2020design}
Jiaxuan You, Zhitao Ying, and Jure Leskovec.
\newblock Design space for graph neural networks.
\newblock \emph{Advances in Neural Information Processing Systems}, 33, 2020.

\bibitem[Zhang \& Chen(2017)Zhang and Chen]{Zhang2017}
Muhan Zhang and Yixin Chen.
\newblock Weisfeiler-lehman neural machine for link prediction.
\newblock In \emph{Proceedings of the 23rd ACM SIGKDD International Conference
  on Knowledge Discovery and Data Mining}, KDD '17, pp.\  575–583, New York,
  NY, USA, 2017. Association for Computing Machinery.
\newblock ISBN 9781450348874.
\newblock \doi{10.1145/3097983.3097996}.
\newblock URL \url{https://doi.org/10.1145/3097983.3097996}.

\bibitem[Zhang \& Chen(2018)Zhang and Chen]{Zhang2018b}
Muhan Zhang and Yixin Chen.
\newblock Link prediction based on graph neural networks.
\newblock In \emph{Advances in Neural Information Processing Systems}, pp.\
  5165--5175, 2018.

\bibitem[Zhou et~al.(2020)Zhou, Cui, Hu, Zhang, Yang, Liu, Wang, Li, and
  Sun]{Zhou2020}
Jie Zhou, Ganqu Cui, Shengding Hu, Zhengyan Zhang, Cheng Yang, Zhiyuan Liu,
  Lifeng Wang, Changcheng Li, and Maosong Sun.
\newblock Graph neural networks: A review of methods and applications.
\newblock \emph{AI Open}, 1:\penalty0 57--81, 2020.
\newblock ISSN 2666-6510.
\newblock \doi{https://doi.org/10.1016/j.aiopen.2021.01.001}.
\newblock URL
  \url{https://www.sciencedirect.com/science/article/pii/S2666651021000012}.

\bibitem[Zhou et~al.(2019)Zhou, Song, Huang, and Hu]{zhou2019auto}
Kaixiong Zhou, Qingquan Song, Xiao Huang, and Xia Hu.
\newblock Auto-gnn: Neural architecture search of graph neural networks.
\newblock \emph{arXiv preprint arXiv:1909.03184}, 2019.

\end{thebibliography}
\bibliographystyle{iclr2022_workshop}

\pagebreak


\appendix

\section{Dataset Details}
\label{dataset-details}

\paragraph{Cora} 
Cora dataset \citep{Sen2008} consists of 2708 scientific publications from Cora and 5429 citation links. Each scientific publication belongs to one of the seven classes and is described by a binary-valued word vector, which indicates the absence/presence of the corresponding words from the dictionary. The dictionary contains 1433 unique words.	
	
\paragraph{CiteSeer} 
CiteSeer dataset \citep{Sen2008} consists of 3312 scientific publications from CiteSeer and 4732 citation links. Each scientific publication belongs to one of the six classes and is described by a binary-valued word vector, which indicates the absence/presence of the corresponding words from the dictionary. The dictionary contains 3703 unique words.

\section{Model Configuration and Experimental Set-up}
\label{model-configuration-and-experimental-set-up}

In this section, we provide details regarding model configuration and experimental set-up from Section \ref{experiments}.

\paragraph{Model configuration}
Experiments were conducted using the following model configuration. Our model consists of $K$ GCN layers applied sequentially, where before each GCN layer there is a dropout of $0.2$, and after each layer but the last there is a ReLU activation. As input it accepts nodes' features $h_v^{(0)} = x_v \in \mathbb{R}^{d}$ and it outputs final hidden representations $h_v = h_v^{(K)} \in \mathbb{R}^{o}$. All intermediate hidden representations have the same dimensionality as the final representation: $h_v^{(i)} \in \mathbb{R}^{o}$ for $i = 1, 2, \dots, K - 1$. The model predicts the probability of an edge between nodes $u$ and $v$ according to the following equation:

\begin{equation}
    P((u, v) \in E) = \sigma \left( h_u \cdot h_v \right),
    \label{edge-probability}
\end{equation}
\noindent
where $\sigma$ represents sigmoid function and $\cdot$ denotes dot product. In experiments, $d$ depends solely on the dimensionality of the graph's features, while $o$ is a hyperparameter. The model is always trained for $500$ epochs using binary cross-entropy as loss function, however, the quality of the model is rather measured by AUC metric. Its parameters are optimized by stochastic gradient descent (SGD) with the learning rate of $0.1$, the momentum of $0.9$, and the weight decay of $5 \cdot 10^{-4}$. During training, a cosine annealing \citep{Loshinov2017} schedule is used for alternation of the learning rate. 

\paragraph{Experimental set-up}
When comparing AdaGrid with the standard complete search and random search, we are interested principally in absolute performance and the trade-off between training time and performance. Experiments are systematically conducted over various settings including different model configurations, datasets, data split ratios, and negative samplings. We use Cora and CiteSeer datasets. To get more representative results we test the model with two configurations: $K = 2$, $o = 64$ and $K = 3$, $o = 128$. The data is every time divided according to other split ratios: $20/40/40$, $50/25/25$, and $80/10/10$. Models are trained and evaluated using the standard uniform sampling as well as community ratio-based negative sampling, proposed in Section \ref{experiments}. In a community ratio-based negative sampling setting, community detection is performed using the Clauset-Newman-Moore greedy modularity maximization algorithm \citep{Clauset2004}. AdaGrid and complete search both consider only the following set of edge message ratios: $Q = \{0.1, 0.2, \dots, 0.9 \}$. To examine power of AdaGrid, it is assessed with various configurations of adapt epochs and try epochs: $(\alpha, \beta) \in \{10, 50, 100 \} \times \{1, 5, \alpha \}$ and both criteria from Section \ref{adaptive-grid-search-adagrid}. Both criteria utilize smoothing. The selection criterion is considered a hyperparameter of AdaGrid, so we present results for the better of the two. Each experiment is repeated three times to mitigate the effect of randomness, and the average performance is reported.

\section{Complete Results}

In Tables \ref{table-complete-adagrid-uniform-negative-sampling} and \ref{table-complete-adagrid-community-ratio-based-negative-sampling} are complete results for the experiments from Section \ref{experiments}. Since the results with $K=3$ and $o=128$ model configuration are similar to the ones with $K=2$ and $o=64$, we report them only for the latter model configuration.

\begin{table}[h!]
    \caption{AUC in percent for AdaGrid, complete search, and random search with uniform negative sampling evaluation. Model has $K=2$, $o = 64$.}
    \label{table-complete-adagrid-uniform-negative-sampling}
    \centering
    \resizebox{\columnwidth}{!}{
    \begin{tabular}{cccccccc}
        \toprule
        & & \multicolumn{6}{c}{\textbf{Datasets}}           \\
        & & \multicolumn{3}{c}{Cora} & \multicolumn{3}{c}{CiteSeer} \\
        \cmidrule(lr){3-5} \cmidrule(lr){6-8}
        \multicolumn{2}{c}{\textbf{Methods}} & $20/40/40$ & $50/25/25$ & $80/10/10$ & $20/40/40$ & $50/25/25$ & $80/10/10$ \\
        \midrule
        \multicolumn{2}{c}{Complete search} & 94.85 & 96.65 & 97.17 & 95.84 & 97.78 & 98.46 \\
        \multicolumn{2}{c}{Random search} & 94.71 & 96.40 & 97.02 & 95.88 & 97.67 & 98.45 \\
        \midrule
        \multicolumn{2}{c}{\textbf{AdaGrid}} \\
        $\alpha$ & $\beta$  \\
        100 & 1 & 95.03 & 96.89 & 97.54 & 95.96 & 97.82 & 98.69 \\
        100 & 5 & 95.03 & 96.87 & 97.63 & 95.91 & 97.80 & 98.64 \\
        100 & 100 & 94.93 & 96.98 & 97.59 & \textbf{95.97} & 97.79 & 98.67 \\
        50 & 1 & 95.01 & 96.98 & 97.68 & 95.94 & 97.81 & \textbf{98.72} \\
        50 & 5 & 95.05 & 97.04 & 97.75 & 95.87 & 97.81 & 98.68 \\
        50 & 50 & 95.03 & 97.06 & 97.76 & \textbf{95.97} & 97.83 & 98.71 \\
        10 & 1 & 95.01 & 97.07 & \textbf{97.90} & 95.79 & \textbf{97.89} & 98.70 \\
        10 & 5 & \textbf{95.14} & \textbf{97.10} & 97.83 & 95.81 & 97.83 & 98.71 \\
        10 & 10 & 95.10 & 97.09 & 97.86 & 95.94 & 97.88 & 98.71 \\
        \midrule
        \multicolumn{2}{c}{Gain} & 0.29 & 0.45 & 0.73 & 0.09 & 0.11 & 0.26 \\
        \bottomrule
    \end{tabular}
    }
\end{table}

\begin{table}[h!]
  \caption{AUC in percent for AdaGrid, complete search, and random search with community ratio-based negative sampling evaluation. Model has $K=2$, $o = 64$.}
  \label{table-complete-adagrid-community-ratio-based-negative-sampling}
  \centering
  \resizebox{\columnwidth}{!}{
  \begin{tabular}{cccccccc}
    \toprule
    & & \multicolumn{6}{c}{\textbf{Datasets}}           \\
    & & \multicolumn{3}{c}{Cora} & \multicolumn{3}{c}{CiteSeer} \\
    \cmidrule(lr){3-5} \cmidrule(lr){6-8}
    \multicolumn{2}{c}{\textbf{Methods}} & $20/40/40$ & $50/25/25$ & $80/10/10$ & $20/40/40$ & $50/25/25$ & $80/10/10$ \\
    \midrule
    \multicolumn{2}{c}{Complete search} & 84.01 & 82.98 & 84.62 & 83.87 & 82.22 & 83.96  \\
    \multicolumn{2}{c}{Random search} & 84.00 & 82.05 & 83.82 & 83.87 & 82.61 & 83.65 \\
    \midrule
    \multicolumn{2}{c}{\textbf{AdaGrid}} \\
    $\alpha$ & $\beta$  \\
    100 & 1 & 84.35 & 83.05 & 85.27 & \textbf{84.01} & 83.20 & 84.58 \\
    100 & 5 & 84.52 & 82.88 & 85.43 & 83.92 & 82.86 & 84.34 \\
    100 & 100 & 84.49 & 83.61 & 85.57 & 84.00 & 83.12 & 84.75 \\
    50 & 1 & 84.51 & 83.48 & 86.02 & 83.88 & 83.11 & 84.79 \\
    50 & 5 & 84.59 & 83.56 & 85.98 & 83.90 & 82.94 & 84.84 \\
    50 & 50 & 84.59 & 83.74 & 85.74 & 84.00 & 82.99 & 84.81 \\
    10 & 1 & 84.49 & 83.69 & 86.27 & 83.65 & 83.11 & \textbf{84.90} \\
    10 & 5 & \textbf{84.65} & 83.82 & 86.54 & 83.75 & \textbf{83.37} & 84.89 \\
    10 & 10 & 84.59 & \textbf{84.07} & \textbf{86.56} & 83.94 & 83.04 & 84.83 \\
    \midrule
    \multicolumn{2}{c}{Gain} & 0.64 & 1.09 & 1.94 & 0.14 & 0.76 & 0.94 \\
    \bottomrule
  \end{tabular}
  }
\end{table}

\end{document}